# Degenerate neutrality creates evolvable fitness landscapes

James Whitacre[1], Axel Bender[2]

[1]School of Information Technology and Electrical Engineering; University of New South Wales at the Australian Defence Force Academy, Canberra, Australia

[2]Land Operations Division, Defence Science and Technology Organisation; Edinburgh, Australia

*Abstract -* Understanding how systems can be designed to be evolvable is fundamental to research in optimization, evolution, and complex systems science. Many researchers have thus recognized the importance of evolvability, i.e. the ability to find new variants of higher fitness, in the fields of biological evolution and evolutionary computation. Recent studies by Ciliberti et al (Proc. Nat. Acad. Sci., 2007) and Wagner (Proc. R. Soc. B., 2008) propose a potentially important link between the robustness and the evolvability of a system. In particular, it has been suggested that robustness may actually lead to the emergence of evolvability. Here we study two design principles, redundancy and degeneracy, for achieving robustness and we show that they have a dramatically different impact on the evolvability of the system. In particular, purely redundant systems are found to have very little evolvability while systems with degeneracy, i.e. distributed robustness, can be orders of magnitude more evolvable. These results offer insights into the general principles for achieving evolvability and may prove to be an important step forward in the pursuit of evolvable representations in evolutionary computation.

## 1. Introduction

Evolvability describes a system's ability to discover new variants of higher fitness. The importance of evolvability is well recognized by many researchers studying biological evolution [1] [2] [3] [4] [5] and Evolutionary Computation (EC) [6] [7] [8] [9]. By describing natural selection as a process of retaining fitter variants, Darwin implicitly assumed that repeated iterations of variation and selection would result in the successive accumulation of useful variations [3]. However, decades of research applying Darwinian principles to computer models have irrefutably demonstrated that the founding principles of natural selection are an incomplete recipe for evolving systems of unbounded complexity. In computer simulations, adaptive changes (i.e. innovations) are at best finite and at worst short-lived. Understanding the origin of innovations is one of the most important open questions that a theory of evolution must still address [10].

In EC, the study of evolvability has mostly focused on a closely related topic; the searchability of a fitness landscape. Because almost every aspect of the design of an Evolutionary Algorithm (EA) can influence its ability to search a particular fitness landscape, there are a broad number of ways in which this problem has been studied. One approach that will not be discussed here, is the design and implementation of variation operators, which we consider to include also the learning of the epistatic linkage between genes, the development of metamodels of fitness functions, and the discovery of building blocks. Although we neglect these issues here, clearly how variation is imposed on a population does influence evolvability [11] as well as the effectiveness of a search process, e.g. see [12].

Another useful way to study evolvability is within the context of the so-called "representation problem". When designing an EA, it is necessary to represent a problem in parametric form (i.e. the genotype) that is then expressed (as a phenotype) through some mapping process and is finally evaluated for fitness. The challenge is to develop a "good mapping" from genotype to phenotype (G:P mapping) so that the fitness landscape is searchable from the perspective of the search bias ingrained within an EA. The encoding of the genotype (i.e. representation) has been actively studied in the EC community [13] [14] [15] [16] [17] [18] [19] [20] [7] [8] [9]. For a recent book on the subject, see [21].

Studying evolvability as a representation problem allows us to use knowledge of the G:P mapping process in biology to inform studies in EC. A number of EC studies have investigated features of the biological G:P mapping process, such as mechanisms for expressing complex phenotypes from compact genetic representations. This is seen for instance in the study of G:P mappings that incorporate protein expression [13] [14] or that simulate biological growth and development [18] [19]. These approaches seem promising, given the importance of development in the evolvability of a species [22] (but also see [23]).

Recently, evolutionary biology has had a resurgent interest in the role of fitness neutrality in evolution [2] [24] [25] [26] [27] [28] [29] [10] [5]. These developments have been followed by the EC community, and some have started to investigate whether increasing neutrality (e.g. artificially introducing a many-to-one mapping between genotypes and phenotypes) can improve the evolvability of a search process [30] [15] [16] [20] [7] [8] [9]. The most common approach in these studies has been to introduce a basic genetic redundancy [30] [15] [8] [9]. Although some studies have suggested that simple redundant forms of neutrality can improve an EA's evolvability, others have questioned the actual utility of fitness landscape neutrality that is introduced through redundant encodings [16].

A few studies have investigated neutrality more closely and have considered different ways that neutrality can be introduced. In [20], it is suggested that redundancy in the G:P mapping is only useful when the genotypes that map to the same phenotype are genetically similar (i.e. close in genotype space) and when higher fitness phenotypes are overrepresented. In [7], evidence is provided that weak coupling between genes (described as reduced ruggedness in a fitness landscape) is needed for neutrality to enhance evolvability.

This study investigates different forms of neutrality that are inspired by observations of biological systems. In particular, the neutrality is generated through mechanisms for achieving robust phenotypes. Our chief concern is to understand the necessary conditions for evolvability, how these conditions are attained in biological systems, as well as the origins of "useful neutrality" in evolution.

In the next section, we define phenotypic variability and explain why it is an important precondition and useful surrogate measure for evolvability. We then touch upon recent developments that have indicated evolvability might be an emergent property of robust complex systems. We also introduce redundancy and degeneracy as two distinct design principles for achieving robustness and neutrality in biological systems. Section 3 presents a simulation model that is used to investigate how these design concepts influence evolvability. The results in Section 4 point to an important role for degeneracy (and not redundancy) in the emergence of evolvability. A brief discussion and conclusions finish the paper in Sections 5 and 6.

## 2. Robustness and Evolvability

### 2.1 Evolvability

Many different definitions of evolvability exist in the literature (e.g. [6] [5] [11]), so it is important to articulate what we mean when we use this term. In general, evolvability is concerned with the selection of new phenotypes. It requires an ability to generate distinct phenotypes and it requires that some of these phenotypes have a non-negligible probability of being selected by the environment. Given the important role the environment plays in the selection process, studies of biological evolution often consider the ability to generate distinct phenotypes as an important precondition and a useful proxy for evolvability.

Similarly, in this study we use Kirchner and Gerhart's definition, which defines evolvability as "*an organism's capacity to generate heritable phenotypic variation*" [4]. To further clarify the meaning of this definition, it is worth differentiating between phenotypic variation and phenotypic variability (evolvability) [11]. Phenotypic variation is the simultaneous existence of distinct phenotypes (e.g. in a population); i.e. it is a directly measurable property of a set of distinct phenotypes. On the other hand, phenotypic variability is a dispositional concept, namely the potential or the propensity for phenotypic variation. More precisely, it is the total accessibility of distinct phenotypes. As with other studies [1] [2] [5], we thus use phenotypic variability as a proxy for a system's evolvability.

### 2.2 Robustness and Evolvability

Recent studies [2] [5] have indicated that robustness may allow for, or even encourage, evolution in biology. At first this may seem surprising since increasing robustness appears to be in direct conflict with the requirements of evolvability. As illustrated in Figure 1, the conflict comes from the apparently simultaneous requirement to robustly maintain developed phenotypes while continually exploring and finding new ones. For example, species are highly robust to internal and external perturbations while on the other hand, evolution has demonstrated a capacity for continual innovation for billions of years. In spite of this apparent conflict, these studies suggest that increasing robustness can sometimes enhance a system's evolvability [2] [5].

In [2] [5] it was speculated that robustness increases evolvability, largely through the existence of a neutral network that extends far throughout the fitness landscape. On the one hand, robustness is achieved through a connected network of equivalent (or nearly equivalent) phenotypes. Because of this connectivity, we know that some mutations or perturbations will leave the phenotype unchanged [5], the extent of which depending on the local network topology.

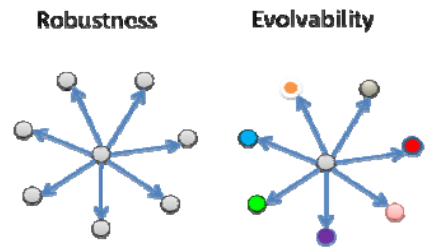

**Figure 1 conflicting forces of robustness and evolvability. A system (central node) is exposed to changing conditions (outer nodes). Robustness of a phenotype requires minimal variation (left) while the discovery of new phenotypes requires exploration of a large number of phenotypic variants (right).**

On the other hand, evolvability is achieved over the long-term by movement across a neutral network that reaches over widely different regions of the fitness landscape. This assumes that different regions of the landscape can access very distinct phenotypes, as was found to occur in the study of artificial gene regulatory networks in [2]. In short, the size and topology of the neutral network could allow evolution to explore a broad range of phenotypes while maintaining core functionalities.

The work by Ciliberti et al in [2] was not the first to highlight the importance of neutral networks in evolution. A neutral theory of molecular evolution was formulated by Kimura [31] and others have studied neutral networks in computer models of biological systems [26]. The novelty of Ciliberti et al's work is the demonstrated expansive range of accessible phenotypes that they believe emerges as a consequence of robust phenotypic expression. This result leads Ciliberti et al to the exciting (but still tentative) conclusion that a causality exists between reduced phenotypic variation (increased robustness) and enhanced phenotypic variability (increased evolvability). In fact, the authors go even further and suggest that robustness alone is not sufficient but that the topology of the neutral network could also matter greatly.

## 2.3 Design principles for achieving robustness

There are two design principles that are believed to play a role in achieving robustness in biological systems; redundancy and distributed robustness [32] [33]. Redundancy is an easily recognizable design principle that is prevalent in both biological and man-made systems. Here, redundancy is used to refer to a redundancy of parts, that is, identical parts that have identical functionality. It is a common feature in engineered systems where redundancy provides a robustness against environmental variations of a very specific type. In particular, redundant parts can be used to replace parts that fail or can be used to augment output when demand for a particular output increases.[1]

Distributed robustness emerges through the actions of multiple dissimilar parts [33] [34]. It is in many ways unexpected because it is only derived in complex systems where heterogeneous components have multiple interactions with each other. In our experiments we demonstrate that distributed robustness can be achieved through degeneracy.

Degeneracy is ubiquitous in biology as evidenced by the numerous examples provided by Edelman and Gally [32]. Degeneracy, sometimes also referred to as partial redundancy, is a term used in biology to refer to conditions where there is a partial overlap in the functions or capabilities of components [32]. In particular, degeneracy refers to conditions where we have structurally distinct components (but also modules and pathways) that can perform similar roles (i.e. are interchangeable) under certain conditions, yet can play distinct roles in others.

## 3. Experimental Setup

This study investigates whether the design principles for achieving neutrality and robustness will impact a system's evolvability in different ways. We use an exploratory abstract model that has been developed to unambiguously distinguish between redundancy and degeneracy concepts and allows us to explore in detail the relationship between these design principles and evolvability. To help ground the work, we present the model within the context of a transportation fleet mix problem. However, we are confident that these robustness design principles and their influence on evolvability is more general and can be related to other contexts including operations, planning, and evolution.

### 3.1 Transportation model

In its simplest form, the transportation fleet model consists of a set of vehicles and is specified by the types of tasks that each vehicle can accomplish. In particular, we define a set of $n$ vehicles and $m$ task types. Vehicles are characterized by a matrix with components $\delta_{ij}$, which take a value of one if vehicle type $i$ is capable of doing task type $j$ and zero otherwise.

The tasks that are allocated to each vehicle define a vehicle's state vector, which is given by $C_i$. The vector components $C_{ij}$ denote the number of tasks of type $j$ that are allocated to a vehicle of type $i$ within the fleet. Without loss of generality, we assume that each state $C_{ij}$ takes a value of zero whenever $\delta_{ij}$ is zero. Over some unspecified period of time, each vehicle is assumed to be able to accomplish at most $\lambda$ tasks, i.e. for each vehicle type $i$, $\sum_{j \epsilon m} C_{ij} \delta_{ij} = \lambda$. Each vehicle is also restricted to only be capable of dealing with two distinct types of tasks (e.g. see Figure 2), i.e. for each vehicle type $i$ we have $\sum_{j \epsilon m} \delta_{ij} = 2$. In the model, the matrix $\delta$ defines the internal changeable components (i.e. genotype) of a given fleet design. In particular, single mutations to the settings of $\delta$ act to replace one vehicle with another vehicle that may have different task capabilities. We elaborate on these mutations in more detail shortly.

A fleet's utilization or phenotype $T^P$ is defined as the fleet's readiness to accomplish particular tasks. We can define each phenotypic trait of a fleet as a vector whose components contain, for each task type, the number of tasks that all of the vehicles in the fleet are ready to accomplish. For a given task type $j$ the trait vector component is $T_j^P = \sum_{i \epsilon n} C_{ij} \delta_{ij}$. The current operating environment $T^E$ for a fleet consists of a set of tasks that need to be accomplished. The fitness $F$ is then defined in (1). As can be seen from this definition, a fleet is penalized for tasks that it is not prepared to accomplish.

---
[1] This definition of redundancy is not identical to other uses in the EC literature. In many papers, the term is used in a more general way to refer to the existence of a many to one G:P mapping.

$$F(T^P) = -\sum_{j \in m} \theta_j \qquad (1)$$

$$\theta_j = \begin{cases} 0, & T_j^P > T_j^E \\ (T_j^P - T_j^E)^2, & else \end{cases}$$

A fleet attempts to satisfy environmental conditions through control over its phenotype, which involves changing the settings of the vehicle states $C$. We implement an ordered asynchronous updating of $C$ where each vehicle conducts a local search and evaluates the changes to fleet fitness resulting from an incremental increase or decrease in the state value of the vehicle. In other words, we reallocate the vehicle to improve its utilization for the vehicle's set of feasible task types. A change in state value is kept if it improves system fitness. Unless stated otherwise, updating component state values is stopped once the fleet fitness converges to a stable fitness value.[2]

Degeneracy and redundancy are modeled by constraining the setting of the matrix $\delta$, which acts to control how the capabilities of vehicles are able to overlap. In the purely redundant model, vehicles are placed into subsets in which all vehicles are genetically identical. In other words, vehicles within a subset can only influence the same set of traits (but are free to take on distinct state values). In the degenerate model, a vehicle can only have a partial overlap in its capabilities when compared with any other vehicle. A simple illustration of the difference between these two design principles is given in Figure 2.

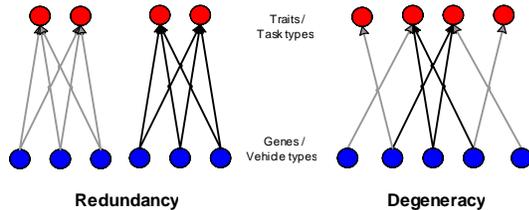

**Figure 2 Illustration of G:P mapping constraints in models of redundancy and degeneracy. A vehicle may be assigned many tasks, however the types of tasks assigned are restricted by the G:P mapping used in a particular fleet.**

### 3.2 Measuring Evolvability

Here we describe the steps used to analyze the evolvability of a fleet, which are similar to those outlined in [2]. The general aim is to discover and analyze, within the fitness landscape, a neutral network and the immediate neighborhood of that network.

First, we consider a network representation of a fitness landscape where each node in the network represents a particular fleet and environment. In our transportation model, this means that a node in the fitness landscape is characterized by a genotype $\delta$, a phenotype $T^P$, and a fitness $F(T^P)$. Connections between nodes represent genetic mutations where an existing vehicle in the fleet is replaced with a different type of vehicle.

A neutral network is then defined as a connected graph of nodes (within the fitness landscape), for which each and every node contains a fleet of the same fitness. Alternatively, one can think of the neutral network as a connected set of genotypes, in which each genotype can reach every other by local movements in genotype space without degrading the system's fitness below that of the stated threshold. In these experiments, we relax the neutrality criteria such that all systems within $\alpha$ % of the optimal fleet fitness are considered to have equivalent fitness. This neutrality relaxation is necessary in order to consider "satisficing behavior" [35]. Justifications for considering an approximate neutrality are varied in the literature, but often are based on constraints observed in physical environments that lead to reductions in selection pressure or limitations to perfect selection.

Similar to [5], we also define a 1-neighborhood representing all non-neutral nodes connected to the neutral network. These nodes correspond to the changes in the fleet that cause a non-neutral change of fitness and that are also reachable from the neutral network. Evolvability (phenotypic variability) of a fleet design is then defined as the total count of unique phenotypes that can be accessed directly from the neutral network (i.e. unique phenotypes within the 1-neighborhood). To understand why this measure of evolvability is meaningful, we elaborate on a possible interpretation of genetic mutations to the fleet.

First, assume that a genetic mutation to the fleet represents the replacement of a vehicle type with a new type of vehicle that is not suitable for any of the currently existing task types. For the purposes of our analysis, this is analogous to a gene deletion or vehicle failure within a fleet. However, it is worth considering what might happen if these new vehicle types could in some rare cases provide an opportunity to achieve new types of tasks that were not conceivable during the fleet design/planning process. Furthermore, assume that the emergence of new task capabilities is dependent on the environment and the system's phenotype.

By allowing for the possibility that these mutations might present new opportunities, the 1-neighborhood obtains new meaning. In particular, although the fleet fails the fitness test from (1), we simply describe these as being non-neutral phenotypes and do not make *a priori* judgments on the fleet's utility. Hence, while we can not directly model innovation, we consider the diversity of these phenotypes a precondition for innovation.

---
[2] This phenotypic control might also be interpreted as local search or Baldwinian evolution, depending on the context.

*3.2.1 Fitness landscape exploration*

Measurement of evolvability requires an exploration of both the neutral network and the 1-neighborhood. Starting with an initial fleet and a given external environment, defined as the first node in the neutral network, the neutral network and 1-neighborhood are explored by iterating the following steps: 1) select a node from the neutral network at random; 2) mutate the fleet; 3) allow the fleet to modify its phenotype in order to adapt to the new conditions; and 4) if fitness is within $\alpha$ % of initial fleet fitness then the fleet is added to the neutral network, else it is added to the 1-neighborhood.

Additions to the neutral network and 1-neighborhood must represent unique genotypes, meaning that duplicate genotypes are discarded when encountered by the search process. The size of the neutral network and 1-neighborhood are too large to allow for an exhaustive search and so the neutral network search algorithm includes a stopping criteria of 20,000 steps (genetic changes).

**Remaining conditions:** Unless stated otherwise, the following experimental conditions are observed in all experiments. Vehicle state values are randomly initialized as integer values between 0 and 10 and genotypes are randomly initialized but constrained to meet the requirements of degeneracy or redundancy, depending on the model tested. The environment is defined with an optimal phenotype that is identical to the initial fleet phenotype and does not change once initialized. The neutrality threshold is set to $\alpha = 5\%$, the number of task types is set to $n=16$ and the number of vehicles in a fleet is set to $m=2n$. Ad hoc experiments varying the settings of $n$ and $m$ did not significantly alter our results. Results are averaged over 50 runs.

## 4. Results

First, we investigate whether the different design principles influence the size of the neutral network and the evolvability of the fleet. Results are shown in Figure 3 for the search algorithm's exploration of the neutral network and 1-neighborhood. Presenting the results in this way allows one to observe the rate at which new neutral genotypes and non-neutral phenotypes (i.e. the "innovation rate" [36]) are being discovered during the search process.

On average, after 20,000 search steps the degenerate system is found to have a much larger neutral network ($NN_{deg} = 1309$, $NN_{red} = 576$) and is at least 10-times more evolvable. This seems to suggest that even moderate increases in the size of the neutral network can lead to dramatic improvements in evolvability. In order to explore this idea further, we modify the models and introduce additional excess vehicle resources to both fleet designs. In particular, we employ the same experimental conditions used previously, except now we increase the number of vehicles in each fleet ($m$). By increasing the amount of resources while maintaining the same number of task requirements, we are expecting that both types of fleets will become more robust to gene deletions and will hence be able to establish larger neutral networks. Results are reported in Figure 4 as the final calculations for neutral network size and evolvability that are obtained after running the search algorithm for 20,000 genotype changes.

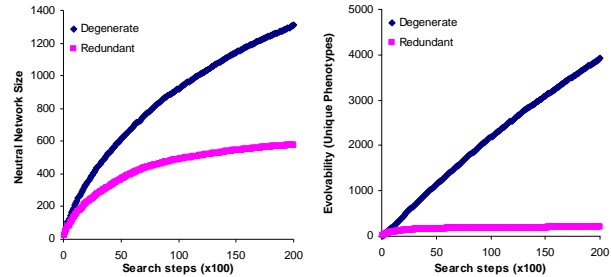

**Figure 3** count of fitness-neutral genotypes (left) unique non-neutral phenotypes (right) discovered during search.

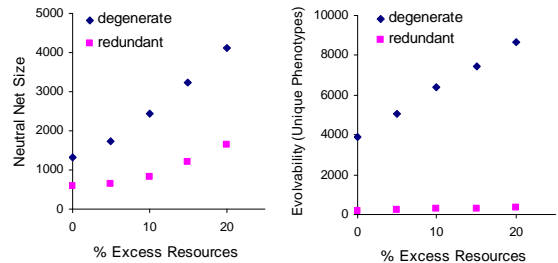

**Figure 4** Estimated neutral network size (left) and evolvability (right) as new redundant and degenerate vehicles are added to the respective fleets.

As indicated in Figure 4, adding excess resources increases the size of the neutral network for both types of fleets. Surprisingly however, the redundant system does not display any substantial increase in evolvability as its neutral network grows. In contrast, the degenerate system is found to have large increases in evolvability, becoming orders of magnitude more evolvable compared with the redundant model, with only modest increases in fleet size. The most important conclusion drawn from these results is that the size of the neutral network within a fitness landscape does not necessarily lead to differences in evolvability, which refutes our earlier speculation. This can be directly observed from the results in Figure 4 by comparing the evolvability of different fleet types for conditions where they have similar neutral network sizes.

In separate experiments (results not shown), we found that changes to the neutrality threshold ($\alpha$) have a similar impact on fleet behavior. Threshold relaxation results in larger neutral networks, however, only for the degenerate system does the evolvability improve markedly.

## 5. Discussion

These experiments provide new insights into the relationship between neutrality and evolvability. While purely redundant encodings are not likely to provide access to distinct phenotypes, degenerate robustness appears to increase phenotypic variability and hence provides the foundation for higher system evolvability. Although these results do not directly investigate evolvable representations in EC, this study provides a theoretical basis for future developments in this area.

Although we have discovered that the design principles for achieving robustness can determine whether a system is evolvable, it is still not clear why exactly this is the case. Some have suggested that the topological properties of the neutral network may partly determine whether robust phenotypic expression leads to evolvability [2]. However, preliminary analysis of the topological properties (e.g. path length, degree average) of the neutral networks studied here have not indicated substantial differences in network topology between the redundant and degenerate models, once network size effects are accounted for.

In this work, we did not directly evaluate the robustness of the different fleet models and so we cannot comment on the actual relationship between robustness and evolvability. Having said that, the fact that the degenerate system can effectively operate under a broader range of genotypic conditions (as evidenced by the neutral network size) suggests that such systems are more robust, at least to this type of change in conditions. Our future work will investigate the robustness afforded by these design principles to determine if the level of system robustness can account for the observed differences in evolvability. Preliminary results indicate that high robustness in redundant systems does not always result in high evolvability. Taken in light of the results presented here, this suggests that it is not simply the size of the neutral network within a fitness landscape or the existence of high robustness that determines the evolvability of a system. Instead it could be the design principles used to achieve robustness and neutrality that matter.

Finally, we should note that the phenotypic expressiveness in our model is bounded within a predefined state space, without any possibility of elaboration. In order to achieve a greater distinctiveness in phenotypic expression, developmental processes directed by a compact genetic representation are almost certainly essential. In biology for instance, the developmental growth of a phenotype and its plasticity in the external environment is critical to the elaboration of more complex expressive forms. Hence, we are not claiming to have completely "solved" the evolvability question as it pertains to natural or artificial evolutionary processes. However, before attempting to tackle these grander challenges in EC and artificial life, it is important to understand how design principles can lead to accessible diversity in phenotypic expression. Here we have shown that degeneracy may play an important role in achieving this precondition of evolvability.

## 6. Conclusions

This study demonstrates that the design principles used to achieve robustness/neutrality in a fitness landscape can dramatically affect the accessibility of distinct phenotypes and hence the evolvability of a system. In agreement with [16], we find that a many-to-one G:P mapping does not guarantee a highly evolvable fitness landscape. However, we also discovered that distributed robustness or degeneracy can result in remarkably high levels of evolvability. Degeneracy is known to be a ubiquitous property of biological systems and is believed to play an important role in achieving robustness [32]. Here we have suggested that the importance of degeneracy could be much greater than previously thought. It actually may act as a key enabling factor in the evolvability of complex systems.